\documentclass[10pt,twocolumn,letterpaper]{article}

\usepackage[pagenumbers]{cvpr} 

\usepackage{graphicx}
\usepackage{amsmath}
\usepackage{amssymb}
\usepackage{booktabs}
\usepackage[table]{xcolor}
\usepackage{pifont}
\usepackage{multirow}
\usepackage{algorithm}
\usepackage{algpseudocode}
\usepackage{algorithmicx}

\definecolor{cvprblue}{rgb}{0.21,0.49,0.74}
\usepackage[pagebackref,breaklinks,colorlinks,allcolors=cvprblue]{hyperref}

\title{StableTrack: Stabilizing Multi-Object Tracking on Low-Frequency Detections}

\author{
    Matvei Shelukhan\textsuperscript{1*}
    \quad 
    Timur Mamedov\textsuperscript{1,2*}
    \quad 
    Karina Kvanchiani\textsuperscript{1}
    \\
    \hspace{-7pt}\textsuperscript{1}Tevian, Moscow, Russia
    \qquad 
    \textsuperscript{2}Lomonosov Moscow State University
    \\
    \texttt{\small matvei.shelukhan@tevian.ai}
    \quad 
    \texttt{\small me@timmzak.com}
    \quad 
    \texttt{\small karinakvanciani@gmail.com}
}

\begin{document}
\maketitle
\let\thefootnote\relax\footnotetext{\textsuperscript{*} These authors contributed equally.}
\begin{abstract}
    Multi-object tracking (MOT) is one of the most challenging tasks in computer vision, where it is important to correctly detect objects and associate these detections across frames. Current approaches mainly focus on tracking objects in each frame of a video stream, making it almost impossible to run the model under conditions of limited computing resources. To address this issue, we propose StableTrack, a novel approach that stabilizes the quality of tracking on low-frequency detections. Our method introduces a new two-stage matching strategy to improve the cross-frame association between low-frequency detections. We propose a novel Bbox-Based Distance instead of the conventional Mahalanobis distance, which allows us to effectively match objects using the Re-ID model. Furthermore, we integrate visual tracking into the Kalman Filter and the overall tracking pipeline. Our method outperforms current state-of-the-art trackers in the case of low-frequency detections, achieving $\textit{11.6\%}$ HOTA improvement at $\textit{1}$ Hz on MOT17-val, while keeping up with the best approaches on the standard MOT17, MOT20, and DanceTrack benchmarks with full-frequency detections.
\end{abstract}    
\section{Introduction}
\label{sec:intro}
Multi-object tracking (MOT) is a fundamental computer vision task with applications in surveillance~\cite{surveillance,mamedov2022video}, autonomous driving~\cite{autondrive}, and behavior analysis~\cite{peoplecounter,mamedov2021queue}. A primary challenge facing modern tracking systems is their high computational demand. Current methods mainly focus on tracking objects in each frame of a video stream, which limits their deployment. Consequently, there is a practical need for tracking methods capable of operating effectively on low-frequency detections.

End-to-end methods~\cite{motr, trackformer, coltrack} address MOT through single-network architectures that unify detection and tracking. However, following most state-of-the-art methods~\cite{strongsort,tracktrack,hybridsort,deepocsort,ucmc,imprasso}, StableTrack employs the tracking-by-detection (TBD) framework, which decouples the task into object detection and cross-frame association. Specifically, new detections, obtained by detector in the current frame, are matched with existing tracks using a variety of similarity measures to maintain object identities over time.

\begin{figure}[t]
  \centering
    \includegraphics[width=0.9\linewidth]{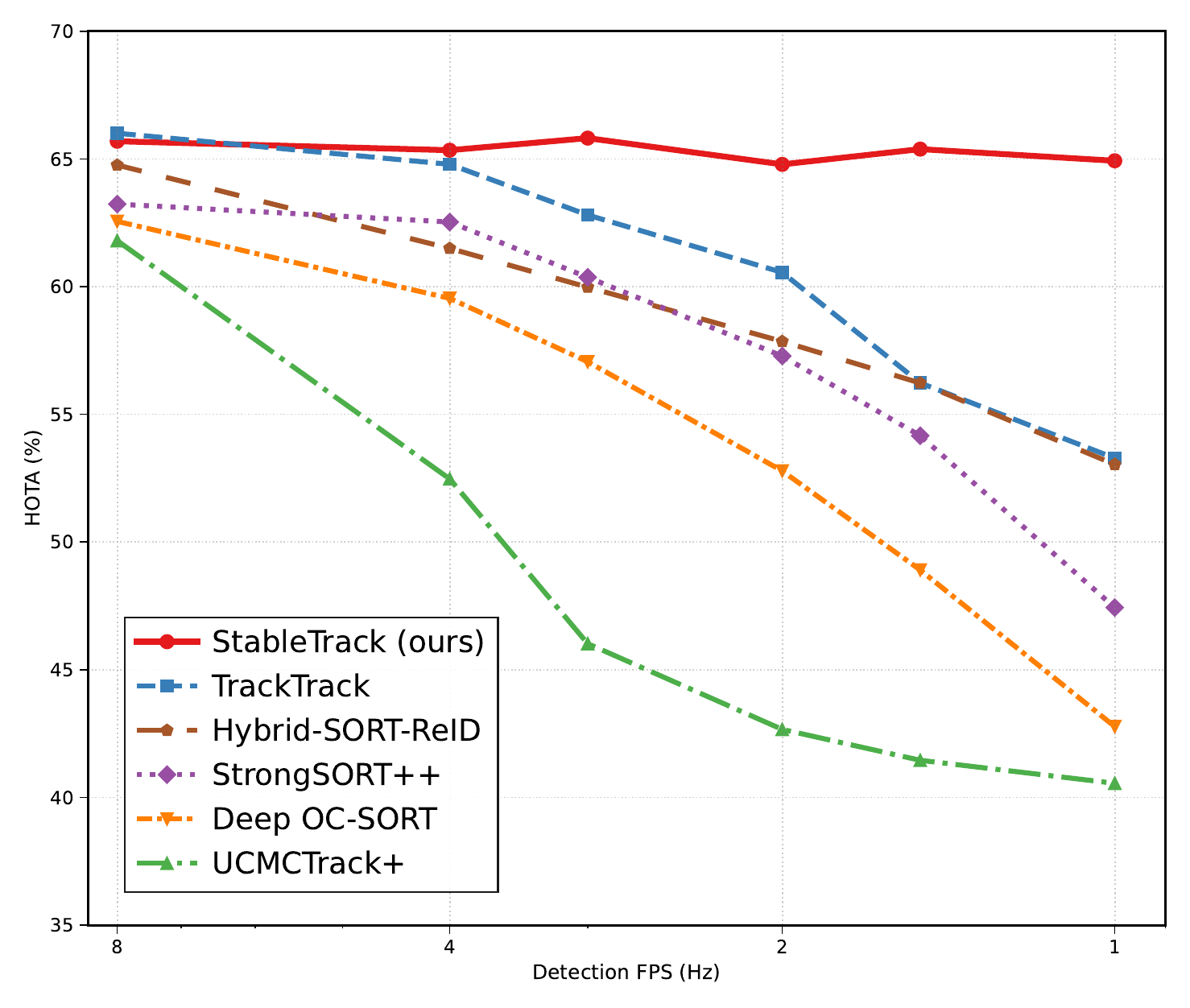}
    \caption{
        Comparison of HOTA scores on the MOT17 validation set with our StableTrack and other state-of-the-art methods for different detection frames per second (FPS). StableTrack shows the highest and stable results in the case of low-frequency detections, while keeping up with the other methods in high-frequency detections scenario.
    }
    \label{fig:fps-comparison}
\end{figure}
Intersection over Union (IoU) between detections and the last bbox of each existing track~\cite{bytetrack, sparsetrack, ocsort} is one of the most common matching metrics. However, direct IoU comparison can fail during temporary occlusions. To address this, most methods~\cite{sort, deepsort, bytetrack} employ a motion model, typically a Kalman Filter (KF)~\cite{kalman}, to predict the position of tracks on the current frame based on their past trajectory. Nevertheless, in the case of low-frequency detections, increased time intervals between detection frames promote motion uncertainty, leading to inaccurate KF predictions. Consequently, reliance on motion cues becomes less effective, increasing the importance of visual appearance similarity.

Re-identification (Re-ID)~\cite{botsort, deepocsort, finetrack, remix} similarity provides a complementary cue based on visual features, which is more robust in low-frequency detection scenarios. However, using Re-ID features alone can result in identity switches in the presence of visually similar objects, as it ignores spatial context. To mitigate this, the Mahalanobis distance~\cite{deepsort, strongsort} is often used in addition to Re-ID similarity to filter improbable associations based on spatial information. This distance metric uses the KF predicted state distribution to assess the likelihood of a detection belonging to a track. However, as mentioned before, KF predictions become less reliable in the case of low-frequency detections, resulting in inaccurate Mahalanobis distance calculations.

To capture both spatial and appearance cues, modern methods commonly fuse multiple similarity measures~\cite{imprasso, hybridsort, tracktrack}, such as IoU, Re-ID, and Mahalanobis distance, into a unified cost matrix. The optimal assignment for the cross-frame association is determined by solving the corresponding cost matrix minimization problem via the Hungarian algorithm~\cite{hungarian}. However, a single matching stage often leaves some tracks and detections unmatched, leading to performance degradation under low-frequency detections.

To address these issues, we propose StableTrack, a novel method designed to stabilize performance on low-frequency detections. Our approach introduces an improved two-stage matching strategy that effectively leverages both visual and spatial information. In the first stage, candidate pairs are filtered using our new Bbox-Based Distance (BBD), adapted for low-frequency detections, before being matched based on Re-ID similarity. This stage allows matching objects based primarily on visual information, thus avoiding KF errors. The second stage processes the remaining unmatched tracks and detections using the Re-ID similarity with a stronger restriction on the spatial similarity of objects, given by the IoU and KF predictions. Additionally, we investigate reliability of KF predictions on low-frequency detections and integrate a visual tracking model to refine KF predictions, thereby enhancing the accuracy of both BBD and IoU calculations for the cross-frame association.

\vspace{2pt}\noindent The main contributions of our work are:
\begin{itemize}
    \item We propose a novel, robust two-stage matching strategy that effectively leverages both visual and spatial information in the case of low-frequency detections.
    \item We introduce a new Bbox-Based Distance (BBD), an adapted for low-frequency detections similarity measure that effectively filters unlikely pairs by jointly considering spatial and temporal information for each track.
    \item We incorporate a visual tracking model to estimate object positions on the intermediate frames, enhancing the KF prediction accuracy and the overall robustness of the cross-frame association in the case of low-frequency detections.
\end{itemize}
This work focuses on human-centric tracking using a person re-identification model, employing the MOT17, MOT20, and DanceTrack datasets for experimental validation. Extensive experiments demonstrate that our method outperforms current state-of-the-art trackers on low-frequency detections while achieving competitive performance on MOT17, MOT20, and DanceTrack benchmarks with full-frequency detections, as illustrated in \cref{fig:fps-comparison}.
\section{Related Work}
\label{sec:rel_work}

\subsection{Cross-Frame Association}

The Hungarian algorithm~\cite{hungarian} remains the cornerstone of the cross-frame association in most modern MOT methods. Numerous techniques have been developed to extend the cross-frame association based on this algorithm. DeepSORT~\cite{deepsort} introduced a matching cascade that prioritizes the association of tracks that have been observed more recently. ByteTrack~\cite{bytetrack} improved cross-frame association completeness by incorporating low-confidence detections through a two-stage matching procedure, first associating high-confidence detections by Re-ID similarity and then low-confidence ones by IoU. OC-SORT~\cite{ocsort} added an additional recovery stage to handle unmatched tracks and detections after the initial cross-frame association. LG-Track~\cite{lgtrack} extended this idea further by proposing a four-stage cross-frame association strategy based on both the localization and classification confidence scores of detections. In a departure from these multi-stage associations, TrackTrack~\cite{tracktrack} abandoned the Hungarian algorithm. Instead, the authors proposed a cross-frame association method that prioritizes local matching precision and incorporates high-confidence detections before the Non-Maximum Suppression (NMS) step, preserving object candidates that would otherwise be discarded.

Although effective in full-frequency benchmarks, these methods struggle with low-frequency detections. To address this, we propose a novel two-stage matching strategy designed to efficiently and accurately match objects across long temporal intervals.

\begin{figure*}
  \centering
  \includegraphics[width=0.9\linewidth]{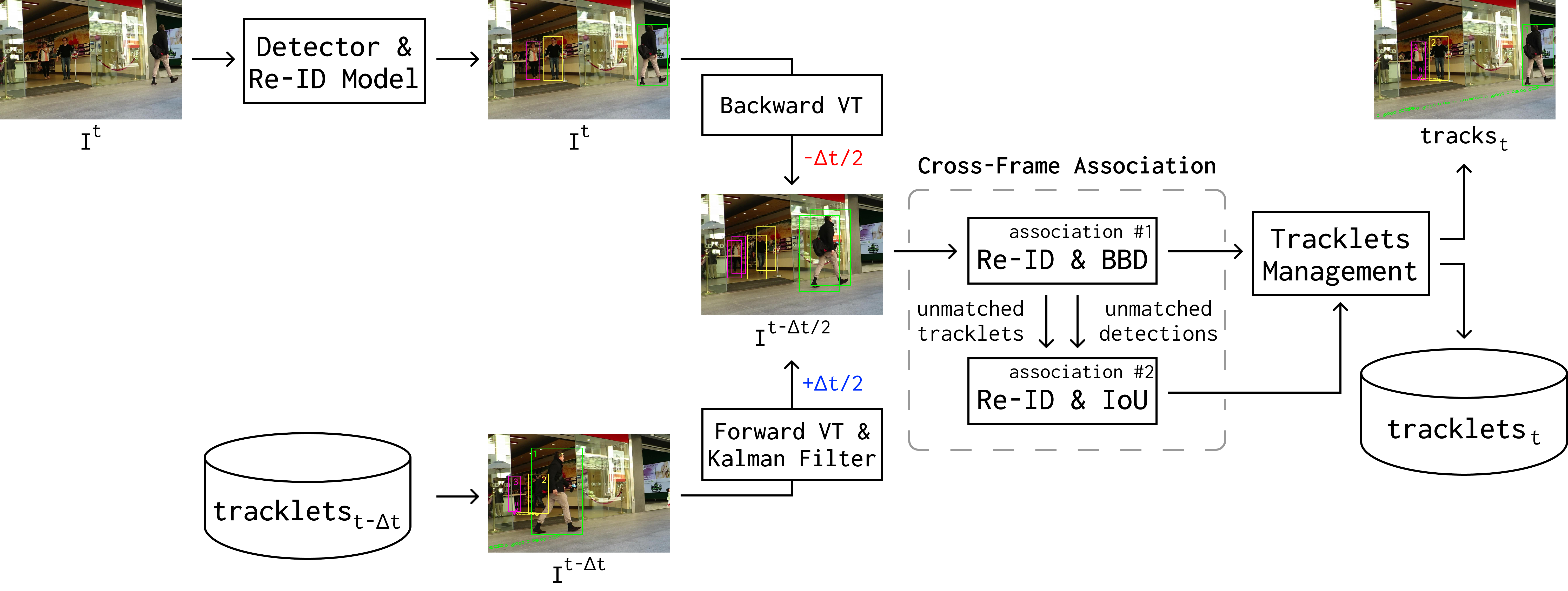}
  \caption{Scheme of StableTrack. The core stages include visual tracking (Forward and Backward VT) for motion modeling, Bbox-Based Distance (BBD) for similarity measuring and two-stage matching strategy for the cross-frame association.}
  \label{fig:pipeline}
\end{figure*}
\subsection{Similarity Measures}

Various similarity measures were employed for the cross-frame association in MOT. The fundamental SORT method~\cite{sort} relied on IoU to associate the Kalman Filter predictions with new detections. However, IoU does not incorporate other spatial characteristics of objects that could improve the robustness of the tracking. Subsequent work has proposed numerous extensions to address this limitation. Generalized-IoU (GIoU)~\cite{giou} extended the original metric to consider the minimum enclosing box for non-overlapping detections. Distance-IoU (DIoU)~\cite{diou} explicitly incorporated the Euclidean distance between the centers of the bounding boxes to improve the cross-frame association of distant objects. Hybrid-SORT~\cite{hybridsort} introduced Height Modulated IoU (HMIoU), upgrading the standard IoU by the similarity of object heights, alongside additional affinity terms based on detection confidence and velocity direction. Rather than building upon IoU, DeepSORT~\cite{deepsort} utilized the Mahalanobis distance as a gating mechanism to filter improbable associations. UCMCTrack~\cite{ucmc} reformulated the tracking problem in world coordinates, performing cross-frame association on the ground plane using a Mapped Mahalanobis Distance.

However, a common drawback of these measures is their susceptibility to performance degradation under low-frequency detections, where large temporal gaps render spatial cues like IoU and Mahalanobis distance less reliable. To address this issue, we propose a novel Bbox-Based Distance (BBD), an adapted for low-frequency detections similarity measure incorporating both spatial and temporal information.
\subsection{Motion Models}
\label{subsec:motionmodels}
Modern tracking methods~\cite{sort, deepsort, tracktrack, bytetrack} employ motion models to predict location of tracks in the current frame, thereby enhancing the robustness of the cross-frame association. The most prevalent class of models assumes Constant Velocity (CV) motion, which presumes objects move with constant velocity between frames. The Kalman Filter~\cite{kalman}, operating under this CV assumption, is widely adopted for this purpose due to its computational efficiency. However, the performance of the Kalman Filter deteriorates under highly dynamic motion or, crucially, on low-frequency detections where its linear predictions become less reliable. To address these limitations, alternative motion modeling approaches have been explored. Some methods~\cite{optflow, ioftracker} leverage optical flow to estimate objects velocity for more accurate short-term prediction. Milan \etal~\cite{Milan_rnn} employed Recurrent Neural Network (RNN) to better predict object trajectories across large temporal gaps. More recently, transformer-based methods~\cite{motr, transmotion, coltrack} have been utilized to model long-range dependencies and contextual motion patterns. Alternatively, the approach based on a Graphs framework~\cite{sushi} was proposed to improve motion modeling, particularly during long-term occlusions.

In most existing approaches, improving the motion modeling has been achieved through the use of complex architectures. Instead, our method uses visual tracking model to improve the cross-frame association between low-frequency detections and refine the Kalman Filter predictions.

\section{Proposed Method}
\label{sec:method}
The scheme of StableTrack is presented in \cref{fig:pipeline}. Let $\mathcal{I} = \{I^1, ..., I^{t - \Delta t}, I^t, I^{t + \Delta t}, ..., I^{\tau}\}$ be a set of frames on which the detector is run, with $\Delta t$ denoting the time interval between consecutive detections. For each frame $I^t \in \mathcal{I}$, the StableTrack produces object tracks and tracklets. A tracklet represents a short-term fragment of an object's trajectory, which may be refined or linked to other fragments in subsequent processing stages. In contrast, a track denotes the final, complete trajectory estimation output by a tracker.

On each frame $I^t \in \mathcal{I}$ in online processing mode, we obtain object detections $\mathcal{D}^t = \{d_1^t, d_2^t, ..., d_m^t\}$ and compute their appearance descriptors using a Re-ID model. The Backward VT module then propagates each detection to the intermediate frame $I^{t - \Delta t/2}$, while the Forward VT module, integrated with Kalman Filter, predicts the state of existing tracklets to the same intermediate frame $I^{t - \Delta t/2}$ (\cref{vt}). This allows us to improve quality of cross-frame association accuracy without increasing computational overhead. The propagated detections and tracklets are then matched through a cross-frame association module (\cref{reidmatch}) utilizing our two-stage matching strategy with Bbox-Based Distance (\cref{bbd}). Finally, the resulting associations are processed by a tracklets management module (\cref{management}). For clarity, we provide the pseudocode for StableTrack in the~\cref{alg:stabletrack}.

\begin{algorithm}
\caption{StableTrack}
\label{alg:stabletrack}
\begin{algorithmic}[1]
\Require Current frame $I^t$, Detection interval $\Delta t$, Existing tracklets $\mathcal{T}^{t - \Delta t}$, Intermediate frame $I^{t-\Delta t / 2}$
\Ensure Set of updated tracklets $\mathcal{T}^t$
\State $\mathcal{D}^t \gets \text{detector}(I^t)$ \Comment{Get all detections}
\State $\mathcal{F}^t \gets \text{reid\_model}(\mathcal{D}^t)$ \Comment{Extract Re-ID features}

\For{each detection $d^t_i \in \mathcal{D}^t$}
    \State $d_i^{t-\Delta t/2} \gets \text{Backward VT}(d^t_i, I^t, I^{t-\Delta t / 2})$
    \State $\mathcal{D} .\text{add}(d_i^{t-\Delta t/2})$
\EndFor

\For{each tracklet $T^{t - \Delta t}_j \in \mathcal{T}^{t - \Delta t}$}
    \State $T_j^{t-\Delta t/2} \gets \text{Forward VT}(T^{t - \Delta t}_j, I^{t-\Delta t}, I^{t-\Delta t / 2})$
    \State $\mathbf{v}_j \gets \text{calculate\_velocity}(T^{t - \Delta t}_j, T_j^{t-\Delta t/2})$
    \State $\widetilde{d}_j^{\;t-\Delta t/2} \gets \text{Kalman Filter}.\text{predict}(T_j^{t-\Delta t}, \mathbf{v}_j)$
    \State $\widetilde{\mathcal{D}} .\text{add}(\widetilde{d}_j^{\;t-\Delta t/2})$
\EndFor

\State $\mathcal{M} \gets \emptyset$ \Comment{Initialize matches set}
\State $\mathcal{M} \gets \text{1'st Association stage}(\mathcal{M}, \widetilde{\mathcal{D}}, \mathcal{D}, \mathcal{F}^t)$
\State $\widetilde{\mathcal{D}}^{'} \gets \widetilde{\mathcal{D}} \backslash \{\widetilde{d}_j|(\widetilde{d}_j, d)\in\mathcal{M}\}$
\State $\mathcal{D}^{'} \gets \mathcal{D} \backslash \{d_i|(\widetilde{d}, d_i)\in\mathcal{M}\}$
\State $\mathcal{M} \gets \text{2'nd Association stage}(\mathcal{M}, \widetilde{\mathcal{D}}^{'}, \mathcal{D}^{'}, \mathcal{F}^t)$

\State \Comment{Tracklets management}
\State $\mathcal{T}^t \gets \text{update\_tracklets}(\mathcal{T}^{t - \Delta t}, \mathcal{M}, \mathcal{F}^t)$
\State $\mathcal{T}^t \gets \text{create\_new\_tracklets}(\mathcal{D}^{t} \backslash \{d^t_i|(\widetilde{d}, d_i)\in\mathcal{M}\})$
\State $\mathcal{T}^t \gets \text{remove\_old\_tracklets}(\mathcal{T}^t)$

\State \Return $\mathcal{T}^t$
\end{algorithmic}
\end{algorithm}
\subsection{Visual Tracking}
\label{vt}
As mentioned in \cref{subsec:motionmodels}, the standard Kalman Filter may be insufficient for accurate object state prediction due to highly nonlinear motion in the case of low-frequency detections. 

To improve the quality of the Kalman Filter, we use a visual tracking model, which is capable to leverage visual information from intermediate frames. The integration of the visual tracking model enables us to predict the bounding box of an object in intermediate frames, thereby enhancing tracking accuracy without additional detector executions, which is the most computationally expensive component in the tracking-by-detection frameworks.  We integrate visual tracking into a Forward and Backward VT modules.

\noindent\textbf{Forward VT.} Let $\mathcal{T}^{t - \Delta t}$ be a set of previously found tracklets. For each tracklet $T_j^{t - \Delta t} \in \mathcal{T}^{t - \Delta t}$ in frame $I^{t - \Delta t}$, the visual tracking model predicts the bounding box of this tracklet in the intermediate frame $I^{t - \Delta t / 2}$ based on the latest origin bounding box of this tracklet. The displacement vector $v_j$ between the centers of the latest origin bounding box and the predicted bounding box is used as a direct velocity observation for the Kalman Filter.

In StableTrack, the observation vector of the Kalman Filter is extended to 6 dimensions $\hat{\mathbf{x}} = [x_c, y_c, w, h, v_j^x, v_j^y]^T$, where $(x_c, y_c)$ are the 2D coordinates of the object center, $(w, h)$ are the width and height, and $(v_j^x, v_j^y)$ is the displacement vector from the visual tracking model. The corresponding modifications of the Kalman Filter parameters are described in more detail in the supplementary materials.

\noindent\textbf{Backward VT.} In addition, we propose a Backward VT module that leverages the visual tracking model to predict the positions of detections in intermediate frames. For each detection $d_i^t \in \mathcal{D}^t$ from the current frame $I^t$, this module predicts the bounding box of this object in the same intermediate frame $I^{t - \Delta t / 2}$.

Subsequently, predicted bounding boxes by the Forward and Backward VT modules are used in the cross-frame association~(\cref{reidmatch}). Due to the integration of Forward and Backward VT modules, the Kalman Filter requires two prediction and two update steps per StableTrack iteration, as detailed in \cref{management}. The Kalman Filter remains essential because it provides a fallback prediction if Forward VT fails to predict object position in the intermediate frame due to occlusion. Furthermore, it smooths the potentially noisy predictions from the visual tracking model. If Backward VT fails to predict object position, the inverse displacement vector calculated by Forward VT or Kalman Filter will be applied as a fallback.

The two-stage VT design mitigates a key limitation of visual trackers: their performance degrades over longer time intervals. We maximize the reliability of visual tracking by limiting the prediction for the Backward VT and Forward VT modules to $\Delta t / 2$, which will be quantitatively analyzed in \cref{ablation}.

\subsection{Bbox-Based Distance}
\label{bbd}
\begin{figure}[t]
  \centering
    \includegraphics[width=0.7\linewidth]{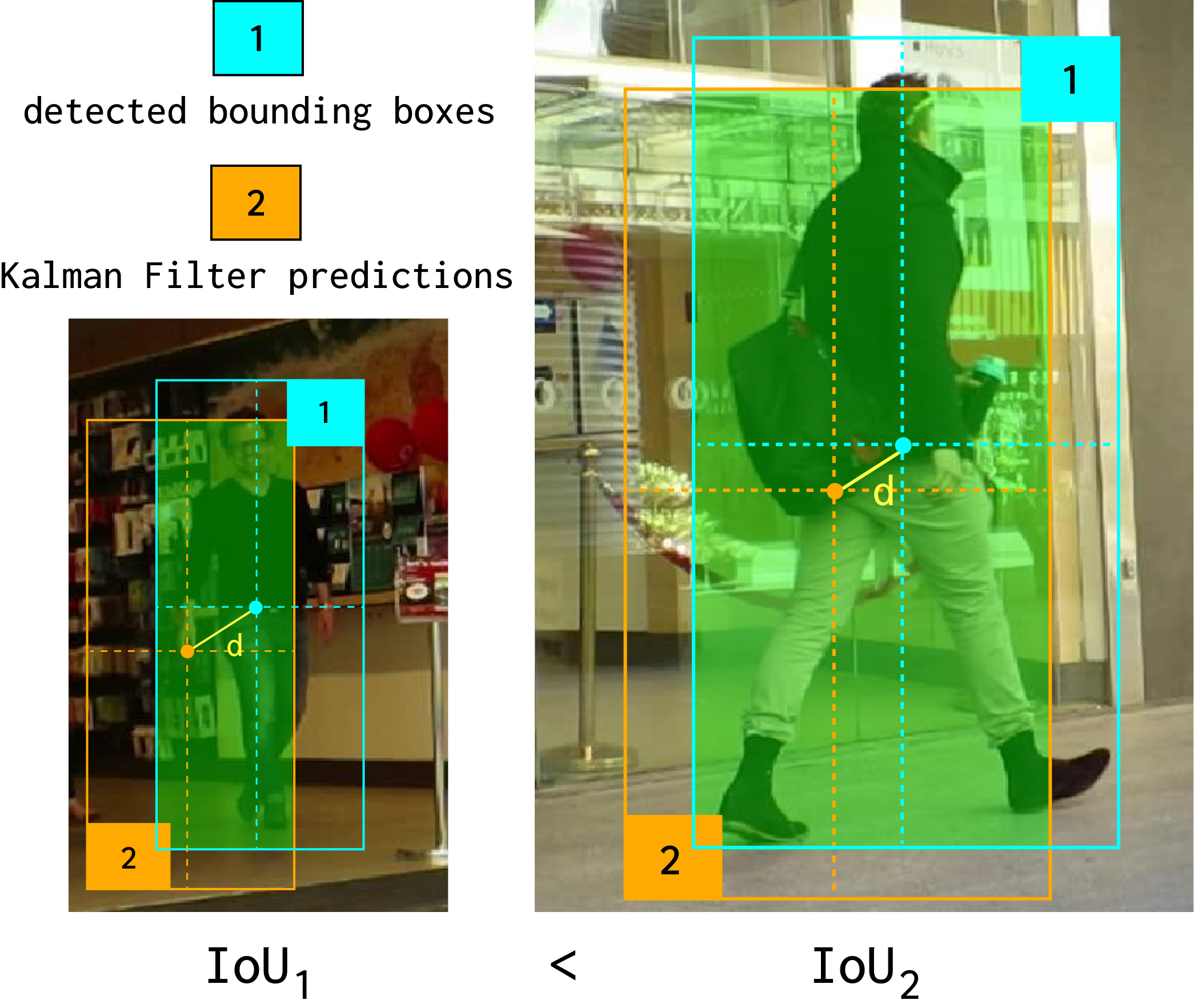}
    \caption{
        The fixed positional displacement $d$ between Kalman Filter prediction and detected bounding box results in different IoU values for smaller objects (left) and for larger ones (right).
    }
    \label{fig:spatial-bbd}
\end{figure}
The Mahalanobis distance is commonly used to filter out improbable associations between Kalman Filter predictions for existing tracklets and detected bounding boxes. Mahalanobis distance computation relies on the Kalman Filter's predicted covariance matrix. However, as established previously, Kalman Filter predictions become less reliable under low-frequency detections, leading to inaccurate covariance estimates. Consequently, the Mahalanobis distance itself becomes an unstable similarity measure for the cross-frame association.

To address this, we propose a novel similarity measure Bbox-Based Distance (BBD), which is an adaptation of the Mahalanobis distance for low-frequency detections. The BBD between a Kalman Filter prediction and detected bounding box is formulated as:

\begin{equation}
D_{\text{BBD}} = \sqrt{(\mathbf{z} - H\hat{\mathbf{x}})^{T}P^{-1}(\mathbf{z} - H\hat{\mathbf{x}})},
\label{eq_bbd}
\end{equation}
where $H$ is the projection matrix that maps the state vector $\hat{\mathbf{x}}$ from Kalman Filter to the image plane, $z = (x, y)$ is the 2D coordinates of the detection's center and $P$ is a covariance matrix that modulates the distance based on spatial and temporal information. Rather than utilizing the estimated covariance matrix from the Kalman Filter in the Mahalanobis distance calculation, we propose a deterministic formulation based on the following expression:
\begin{equation}
P = \begin{bmatrix}
    (c \cdot w)^2 \cdot [\Delta\tau]_\alpha^\beta & 0 \\
    0 & (c \cdot h)^2 \cdot [\Delta\tau]_\alpha^\beta
    \end{bmatrix},
\label{eq_p_matrix}
\end{equation}
where
\begin{align*}
[\Delta\tau]_\alpha^\beta = \min(\max(\Delta \tau, \alpha), \beta),
\end{align*}
$h$ and $w$ represent the height and width from the state vector of Kalman Filter, $\Delta \tau$ is the time interval since the last successful update of the tracklet, $\alpha$, $\beta$ and $c$ are the clipping constants. The design of $P$ incorporates two key intuitions:
\begin{itemize}
    \item \textbf{Spatial Scale}: The influence of the same distance error $d$ exhibits strong dependence on object scale (\cref{fig:spatial-bbd}). To address this scaling dependency, we incorporate the height $h$ and width $w$ parameters into the $P$ matrix.
    \item \textbf{Temporal Uncertainty}: The extended time interval without tracklet updates introduce significant uncertainty in the objects state estimation. To explicitly model this temporal uncertainty, we also integrate the $\Delta \tau$ parameter into the $P$ matrix.
\end{itemize}
Notably, our approach utilizes $h$ and $w$ from the Kalman Filter state vector because, unlike the object center coordinates, these spatial parameters are less affected by low-frequency detections. The Kalman Filter predicts their changes more reliably, making them suitable for our method. The proposed BBD similarity measure is used in cross-frame association (\cref{reidmatch}).

\subsection{Cross-Frame Association}
\label{reidmatch}
During the cross-frame association, new detections, obtained by detector in the current frame $I^t$, are matched with existing tracklets. Our proposed two-stage matching strategy uses both visual and spatial information to correctly match objects across frames by implementing two sequential stages:
\subsubsection{Stage 1: BBD-gated Appearance Matching}
\label{bb_gated_matching}
The first stage employs a cost matrix $C_1$ that combines BBD with Re-ID cosine similarity:
\begin{equation}
C_1(i, j) = 
\begin{cases} 
1 - s_{ij}, & \text{if } 
\begin{aligned}[t]
    &(D_{\text{BBD}}(i, j) < \theta_{\text{BBD}} \;\wedge\; \\
    &\;\;\;\;\;\;\;\;\;\wedge s_{ij} > \theta_{\text{reid-high}})
\end{aligned} \\
\infty      & \text{otherwise},
\end{cases}
\label{c1}
\end{equation}
where 
\begin{align*}
 D_{\text{BBD}}(i, j) = D_{\text{BBD}}(d_i^{t -\Delta t / 2}, \widetilde{d}_j^{\,t - \Delta t / 2})
\end{align*}
is the BBD between the tracklet bounding box $\widetilde{d}_j^{\,t - \Delta t / 2}$, obtained from the Forward VT and Kalman Filter, and detected bounding box $d_i^{t -\Delta t / 2}$, obtained from the Backward VT, $s_{ij}$ is the cosine similarity between the Re-ID embeddings of the tracklet $T_j^{t - \Delta t}$ (\cref{management}) and detection $d_i^{t}$, and $\theta_{\text{BBD}}$ is a threshold on the BBD. The Hungarian algorithm is applied to $C_1$ to find the optimal assignment. Furthermore, any pair assigned with a Re-ID similarity below a threshold $\theta_{\text{reid-high}}$ is rejected to ensure high-confidence matches.

\subsubsection{Stage 2: IoU-gated Appearance Matching}
\label{iou_gated_matching}
The second stage processes all rejected detections and tracklets after \cref{bb_gated_matching}. This stage replaces BBD with IoU filtering to handle cases where the Kalman Filter prediction is accurate, but appearance similarity is lower (e.g., due to partial occlusion). The cost matrix $C_2$ is defined as:
\begin{equation}
C_2(i, j) = 
\begin{cases} 
1 - s_{ij}, & \text{if } 
\begin{aligned}[t]
    &(D_{\text{IoU}}(i, j) < \theta_{\text{IoU}} \;\wedge\; \\
    &\;\;\;\;\;\;\;\;\wedge s_{ij} > \theta_{\text{reid-low}})
\end{aligned} \\
\infty      & \text{otherwise},
\end{cases}
\label{c2}
\end{equation}
where 
\begin{align*}
 D_{\text{IoU}}(i, j) = D_{\text{IoU}}(d_i^{t -\Delta t / 2}, \widetilde{d}_j^{\,t - \Delta t / 2})
\end{align*}
is the IoU between the tracklet bounding box $\widetilde{d}_j^{\,t - \Delta t / 2}$, obtained from the Forward VT and Kalman Filter, and detected bounding box $d_i^{t -\Delta t / 2}$, obtained from the Backward VT. The Hungarian algorithm is applied again to $C_2$, but with a lower Re-ID similarity threshold $\theta_{\text{reid-low}}$ ($\theta_{\text{reid-low}} < \theta_{\text{reid-high}}$).
\begin{table*}[t]
\centering
\begin{subtable}[b]{\linewidth}
\centering
\small
\fontsize{8pt}{10pt}\selectfont
\setlength{\tabcolsep}{2pt}
\renewcommand{\arraystretch}{1.0}
\begin{tabular}{l|cc|cc|cc|cc|cc|cc|cc|cc|cc}
\hline
\noalign{\smallskip}
& \multicolumn{6}{c|}{\textbf{MOT17-val}} & \multicolumn{6}{c|}{\textbf{MOT20-val}} & \multicolumn{6}{c}{\textbf{DanceTrack-val}} \\
\cline{2-19}
 & \multicolumn{2}{c|}{$1$ Hz} & \multicolumn{2}{c|}{$2$ Hz} & \multicolumn{2}{c|}{$4$ Hz} & \multicolumn{2}{c|}{$1$ Hz} & \multicolumn{2}{c|}{$2$ Hz} & \multicolumn{2}{c|}{$4$ Hz} & \multicolumn{2}{c|}{$1$ Hz} & \multicolumn{2}{c|}{$2$ Hz} & \multicolumn{2}{c}{$4$ Hz} \\
\cline{2-19}
Tracker & HOTA & AssA & HOTA & AssA & HOTA & AssA & HOTA & AssA & HOTA & AssA & HOTA & AssA & HOTA & AssA & HOTA & AssA & HOTA & AssA \\
\noalign{\smallskip}
\cline{1-19}
\noalign{\smallskip}
UCMCTrack++ \cite{ucmc} 
& $40.6$ & $32.5$ & $42.7$ & $32.6$ & $52.5$ & $46.0$ & $15.4$ & $4.9$ & $16.9$ & $5.0$ & $35.1$ & $19.5$ & $30.0$ & $11.7$ & $30.1$ & $12.2$ & $30.5$ & $13.4$ \\
Deep OC-SORT \cite{deepocsort} 
& $42.8$ & $46.5$ & $52.8$ & $57.3$ & $59.5$ & $63.9$ & $31.0$ & $31.8$ & $56.9$ & $57.6$ & $64.0$ & $65.1$ & $28.5$ & $16.2$ & $35.9$ & $21.2$ & $45.5$ & $29.1$ \\
StrongSORT++ \cite{strongsort} 
& $47.4$ & $52.9$ & $57.3$ & $63.2$ & $62.5$ & $66.5$ & $\underline{37.4}$ & $\underline{37.5}$ & $\underline{58.6}$ & $\underline{59.4}$ & $64.4$ & $\mathbf{65.5}$ & $27.5$ & $14.3$ & $32.3$ & $17.2$ & $40.4$ & $23.1$ \\
Hybrid-SORT-ReID \cite{hybridsort} 
& $53.0$ & $60.4$ & $57.9$ & $63.4$ & $61.5$ & $64.4$ & $25.5$ & $30.0$ & $53.6$ & $53.4$ & $\underline{64.7}$ & $\underline{65.3}$ & $33.5$ & $\underline{20.7}$ & $42.6$ & $\underline{27.5}$ & $\underline{54.3}$ & $\underline{38.8}$ \\
TrackTrack \cite{tracktrack} 
& $\underline{53.3}$ & $\underline{60.9}$ & $\underline{60.6}$ & $\underline{66.8}$ & $\underline{64.8}$ & $\mathbf{69.9}$ & $34.2$ & $34.7$ & $54.9$ & $53.1$ & $64.4$ & $64.9$ & $\underline{35.9}$ & $18.9$ & $\underline{43.1}$ & $25.3$ & $53.1$ & $36.1$ \\
\hline
\rowcolor{gray!10}
StableTrack (Ours) 
& $\mathbf{64.9}$ & $\mathbf{68.2}$ & $\mathbf{64.8}$ & $\mathbf{67.8}$ & $\mathbf{65.3}$ & $\underline{68.3}$ & $\mathbf{63.5}$ & $\mathbf{61.5}$ & $\mathbf{64.7}$ & $\mathbf{63.4}$ & $\mathbf{65.4}$ & $64.8$ & $\mathbf{56.6}$ & $\mathbf{40.2}$ & $\mathbf{59.0}$ & $\mathbf{43.6}$ & $\mathbf{59.4}$ & $\mathbf{44.3}$ \\
\hline
\end{tabular}
\caption{MOT17-val, MOT20-val and DanceTrack-val sets with low-frequency detections.}
\label{table:low-frequency-results}
\end{subtable}

\begin{subtable}[b]{\linewidth}
\centering
\small
\fontsize{8pt}{10pt}\selectfont
\setlength{\tabcolsep}{2pt}
\renewcommand{\arraystretch}{1.0}
\begin{tabular}{l|cccc|cccc|cccc}
\hline
\noalign{\smallskip}
& \multicolumn{4}{c|}{\textbf{MOT17}} & \multicolumn{4}{c|}{\textbf{MOT20}} & \multicolumn{4}{c}{\textbf{DanceTrack}} \\
\cline{2-13}
Tracker & HOTA & IDF1 & MOTA & AssA & HOTA & IDF1 & MOTA & AssA & HOTA & IDF1 & MOTA & AssA \\
\noalign{\smallskip}
\cline{1-13}
\noalign{\smallskip}
Hybrid-SORT-ReID  \cite{hybridsort}& $64.0$ & $78.7$ & $79.9$ & $63.5$ & $63.9$ & $78.4$ & $76.7$ & $64.5$ & $65.7$ & $67.4$ & $91.8$ & $-$\\
FineTrack  \cite{finetrack}& $64.3$ & $79.5$ & $80.0$ & $64.5$ & $63.6$ & $79.0$ & $77.9$ & $63.8$ & $52.7$ & $59.8$ & $89.9$ & $38.5$ \\
StrongSORT++   \cite{strongsort}& $64.4$ & $79.5$ & $79.6$ & $64.4$ & $62.6$ & $77.0$ & $73.8$ & $64.0$ & $55.6$ & $55.2$ & $91.1$ & $38.6$\\
Deep OC-SORT   \cite{deepocsort}& $64.9$ & $80.6$ & $79.4$ & $65.9$ & $63.9$ & $79.2$ & $75.6$ & $65.7$ & $61.3$ & $61.5$ & $92.3$ & $45.8$ \\
SparseTrack  \cite{sparsetrack}& $65.1$ & $80.1$ & $81.0$ & $65.1$ & $63.4$ & $77.3$ & $78.2$ & $62.8$ & $55.5$ & $58.3$ & $91.3$ & $39.1$\\
CMTrack \cite{cmtrack}& $65.5$ & $81.5$ & $80.7$ & $66.1$ & $64.8$ & $79.9$ & $76.2$ & $66.7$ & $61.8$ & $63.3$ & $\underline{92.5}$ & $46.4$ \\
UCMCTrack++  \cite{ucmc}& $65.7$ & $81.0$ & $80.6$ & $66.4$ & $62.8$ & $77.4$ & $75.6$ & $63.5$
& $63.6$ & $65.0$ & $88.9$ & $51.3$ \\
PIA  \cite{pia}& $66.0$ & $81.1$ & $\underline{82.2}$ & $65.8$ & $64.7$ & $79.0$ & $\underline{78.5}$ & $64.9$ & $-$ & $-$ & $-$ & $-$\\
ImprAsso \cite{imprasso}&  $66.4$ & $82.1$ & $\mathbf{82.2}$ & $66.6$ & $64.6$ & $78.8$ & $\mathbf{78.6}$ & $64.6$ & $-$ & $-$ & $-$ & $-$ \\
TrackTrack \cite{tracktrack}&  $\underline{67.1}$ & $\mathbf{83.1}$ & $81.8$ & $\mathbf{68.2}$ & $\mathbf{65.7}$ & $\mathbf{80.9}$ & $78.0$ & $\mathbf{67.3}$ & $\mathbf{66.5}$ & $\underline{67.8}$ & $\mathbf{93.6}$ & $\underline{52.9}$\\
\hline
\rowcolor{gray!10} StableTrack (Ours)  & $\mathbf{67.1}$ & $\underline{83.0}$ & $82.0$ & $\underline{68.0}$ & $\underline{65.1}$ & $\underline{80.2}$ & $76.9$ & $\underline{66.8}$ & $\underline{66.3}$ & $\mathbf{67.8}$ & $92.0$ & $\mathbf{53.0}$\\
\hline
\end{tabular}
\caption{MOT17, MOT20 and DanceTrack test sets with full-frequency detections.}
\label{table:full-frequency-results}
\end{subtable}
\caption{Comparison of our StableTrack with state-of-the-art methods in the different scenarios. The highest performance metrics are emphasized in boldface, while secondary rankings are denoted with underlining.}
\label{tab:comparison}
\end{table*}
This two-stage matching strategy is designed to address two distinct scenarios. The first stage prioritizes high visual similarity, tolerating larger spatial uncertainty via BBD. The second stage captures pairs where the predicted and detected bounding boxes are well-aligned with high IoU but whose appearance features may be less reliable due to factors like partial occlusion. Since IoU decays more rapidly with distance than BBD, it imposes a stricter spatial constraint, justifying the use of a lower appearance confidence threshold $\theta_{\text{reid-low}}$. It is also worth noting that Forward and Backward VT especially facilitate accurate IoU calculation by propagating both tracklets and detections into an intermediate frame.
\subsection{Tracklets Management}
\label{management}
Following the cross-frame association, the Tracklets Management module performs several critical operations. For tracklets successfully matched with some detections, the Kalman Filter state is updated using temporally predicted detections obtained by the Backward VT module. The Kalman Filter subsequently executes an additional prediction step to propagate tracklets forward to the current frame $I^t$, utilizing inverse velocity estimates derived from the backward VT module. After that, the Kalman Filter state is updated using the original detections from $\mathcal{D}^t$. Re-ID descriptors for matched tracklets are refined via an Exponential Moving Average (EMA), consistent with established practices~\cite{botsort, strongsort}. Previously unmatched detections initiate new tracklets, while tracklets that do not associate over $t_{\text{live}}$ seconds are terminated to mitigate identity accumulation.

\section{Experiments}
\label{sec:exps}
\subsection{Datasets}
We evaluate the proposed method on three challenging and conventional multi-object tracking benchmarks MOT17~\cite{mot17}, MOT20~\cite{mot20} and DanceTrack~\cite{dance}. MOT17 focuses on pedestrian tracking in moderately crowded scenes, featuring both static and moving camera sequences. MOT20 presents highly dense crowds with frequent occlusions, primarily captured by static cameras. DanceTrack comprises multiple video sequences containing people with nearly identical clothing, creating substantial difficulties for re-identification methods. For ablation studies on MOT17, we follow the protocol established in~\cite{tracktrack}, using its four training videos for training and the remaining three training videos for validation. The similar hold-out validation strategy is applied to the MOT20 training set, where we use two training videos for training and the remaining two training videos for validation.

\subsection{Metrics}
We employ standard MOT metrics to comprehensively evaluate tracking performance. MOTA, which is included in the CLEAR~\cite{clear} metric group, primarily assesses detection accuracy. IDF1~\cite{idf1} measures the correctness of long-term identity associations. HOTA~\cite{hota} is designed as a balanced metric, combining detection accuracy (DetA), association accuracy (AssA), and localization accuracy (LocA) into a single unified score. Following recent trends in MOT research~\cite{tracktrack, hybridsort, ucmc, imprasso}, we consider HOTA as our primary metric for optimization and comparison, as it provides the most holistic assessment of tracker performance.

\subsection{Implementation Details}
\label{subsec:impl}
Our framework is implemented in PyTorch. All experiments are conducted on a desktop with an Intel Core i5-10600K CPU and an NVIDIA GeForce RTX 2060 GPU. We use YOLOX-X model~\cite{yolox} as the detector, same as in~\cite{tracktrack, hybridsort, bytetrack, strongsort}. For our feature extractor, we utilize the generalizable Re-ID model from~\cite{dynamix}. The ASMS model~\cite{asms} is used as the visual tracking model due to its simplicity and low computational cost. To ensure a fair comparison in the low-frequency setting, we disable camera motion compensation~\cite{botsort} and specific post-processing techniques~\cite{strongsort, nct} for all evaluated methods, including our own. These components are only activated in the full-frequency detection scenario to align with standard evaluation practices~\cite{tracktrack, botsort, ocsort, strongsort, adaptrack}. Also, Backward VT and Forward VT modules are applied only in the low-frequency protocol. A detailed description of hyperparameters and their selection process is provided in the supplementary material.

\subsection{Comparison with State-of-the-Art Methods}
We compare our method with state-of-the-art trackers under two distinct protocols:
\begin{itemize}
\item Low-Frequency Protocol: A more realistic scenario where computational constraints limit the detector and Re-ID model to run only at a specific frame rate. Since the standard MOT17, MOT20, and DanceTrack benchmarks do not support this protocol, we use their official validation sets for this evaluation. In contrast to existing state-of-the-art trackers, our method is specifically designed for stable performance across both full-frequency and low-frequency scenarios. Consequently, comparative evaluations employ base configurations of other trackers to ensure a fair comparison. Furthermore, within this evaluation protocol, comparative analysis was restricted to methods with publicly available implementations.
\item Full-Frequency Protocol: The standard evaluation scheme in which all methods process every frame of the video sequence. The evaluation results under this protocol are obtained on a hidden test set as part of the official MOTChallenge benchmarking framework.
\end{itemize}

\vspace{2pt}\noindent\textbf{Low-Frequency Detections}: The core strength of our method is evident in the low-frequency protocol. As summarized in \cref{table:low-frequency-results}, StableTrack significantly outperforms all state-of-the-art methods across detection rates of $1$ Hz, $2$~Hz and $4$ Hz on MOT17-val, MOT20-val and DanceTrack-val, leading in almost all metrics. This demonstrates a superior resilience on low-frequency detections. The advantage is most pronounced at the extreme rate of 1 Hz, where our method achieves a notable improvement of $11.6\%$, $26.1\%$ and $20.7\%$ in the primary HOTA metric on MOT17-val, MOT20-val and DanceTrack-val, respectively. In particular, the comparative results demonstrate that our method maintains consistent stable performance across varying detection frequencies.

\vspace{2pt}\noindent\textbf{Full-Frequency Detections}: As shown in \cref{table:full-frequency-results}, our method demonstrates a performance on par with the current state-of-the-art online trackers in MOT17, MOT20 and DanceTrack under the full-frequency protocol. These results underscore the general robustness of our approach, despite it being specifically designed for low-frequency scenarios.

\subsection{Ablation Study}
\label{ablation}
\subsubsection{Component Ablation}
\begin{table}
\centering
\small
\fontsize{8pt}{10pt}\selectfont
\setlength{\tabcolsep}{2pt}
\renewcommand{\arraystretch}{1.0}
\begin{tabular}{ll|cc|cc}
\hline
\noalign{\smallskip}
& & \multicolumn{2}{c|}{\textbf{MOT17-val}} & \multicolumn{2}{c}{\textbf{MOT20-val}} \\
\cline{1-6}
BBD & TSM & HOTA & AssA & HOTA & AssA \\
\noalign{\smallskip}
\cline{1-6}
\noalign{\smallskip}
& & $66.3$ & $67.7$ & $64.7$ & $62.6$ \\
\multicolumn{1}{c}{\ding{51}} & & $66.5$ & $68.1$ & $65.2$ & $63.5$ \\
& \multicolumn{1}{c|}{\ding{51}}  & $69.4$ & $73.3$ & $67.5$ & $67.8$ \\
\hline
\rowcolor{gray!10}
\multicolumn{1}{c}{\ding{51}} & \multicolumn{1}{c|}{\ding{51}} & $\mathbf{69.6}$ & $\mathbf{73.7}$ & $\mathbf{68.1}$ & $\mathbf{69.0}$ \\
\hline
\end{tabular}
\caption{Ablation study for our Bbox-Based Distance (BDD) and  two-stage matching (TSM) in case of full-frequency detections.}
\label{table:abl-main}
\end{table}

\begin{table}
\centering
\small
\fontsize{8pt}{10pt}\selectfont
\setlength{\tabcolsep}{2pt}
\renewcommand{\arraystretch}{1.0}
\begin{tabular}{lll|cc|cc}
\hline
\noalign{\smallskip}
& & & \multicolumn{2}{c|}{\textbf{MOT17-val}} & \multicolumn{2}{c}{\textbf{MOT20-val}} \\
\cline{1-7}
BBD & TSM & VT & HOTA & AssA & HOTA & AssA \\
\noalign{\smallskip}
\cline{1-7}
\noalign{\smallskip}
& & & $63.4$ & $65.1$ & $39.9$ & $24.6$ \\
\multicolumn{1}{c}{\ding{51}} & & & $63.8$ & $65.9$ & $52.3$ & $41.7$ \\
& \multicolumn{1}{c}{\ding{51}} & & $64.1$ & $66.5$ & $40.2$ & $24.9$ \\
\multicolumn{1}{c}{\ding{51}} & \multicolumn{1}{c}{\ding{51}} & & $64.6$ & $67.6$ & $63.2$ & $60.8$ \\
\hline
\rowcolor{gray!10}
\multicolumn{1}{c}{\ding{51}} & \multicolumn{1}{c}{\ding{51}} & \multicolumn{1}{c|}{\ding{51}} & $\mathbf{64.9}$ & $\mathbf{68.2}$ & $\mathbf{63.5}$ & $\mathbf{61.5}$ \\
\hline
\end{tabular}
\caption{Ablation study for our Bbox-Based Distance (BDD), two-stage matching (TSM) and visual tracking (VT) in case of low-frequency detections (1 Hz).}
\label{table:abl-low-fps}
\end{table}
We conduct a comprehensive ablation study to evaluate the contribution of each proposed component. The results are summarized in \cref{table:abl-main} and \cref{table:abl-low-fps}. For both protocols, we establish a baseline that performs a simple one-stage association using only Re-ID similarity without any additional similarity measures~(first row in each table).

\vspace{2pt}\noindent\textbf{Full-Frequency Detections.} In this study (\cref{table:abl-main}), we demonstrate the effectiveness of the proposed components in full-frequency detection scenario. BBD~(second row) improves tracking quality by reducing the candidate matching space, effectively suppressing Re-ID errors. Also, a significant gain is achieved by incorporating TSM~(third row), underscoring the importance of the additional cross-frame association stage even without BBD. The combined application of BBD and TSM~(last row) results in the highest performance. We do not include visual tracking in this study, as there is no reason to deploy visual tracking modules in full-frequency scenarios. 

\vspace{2pt}\noindent\textbf{Low-Frequency Detections.} In this study (\cref{table:abl-low-fps}), we demonstrate the effectiveness of the proposed components in low-frequency detection scenario. The integration of BBD~(second row) yields substantial improvements in tracking performance. This result emphasizes the dual necessities of incorporating spatial scale, since identical Kalman Filter estimation errors produce divergent effects across size-varying objects~(\cref{fig:spatial-bbd}), and temporal uncertainty modeling to address state degradation during update delays. Employing a two-stage matching strategy~(third row) also enhances tracking performance. Critically, the incorporation of BBD within the first stage of the TSM (fourth row) substantially enhances the quality of the cross-frame association, as BBD serves as a fundamental component of the TSM module. The combined application of BBD, TSM, and VT modules~(last row) results in the highest performance, confirming the importance of using visual information from intermediate frames for reliable trajectory estimation. Furthermore, VT is used to compute both IoU and BBD within the intermediate frame, so there is no reason to investigate the VT impact in isolation from BBD and TSM.

\subsubsection{Bbox-Based Distance vs. Mahalanobis Distance}
\begin{table}
\centering
\small
\fontsize{8pt}{11pt}\selectfont
\setlength{\tabcolsep}{0.8pt}
\renewcommand{\arraystretch}{1.0}
\begin{tabular}{l|cc|cc|cc|cc}
\hline
\noalign{\smallskip}
& \multicolumn{4}{c|}{\textbf{MOT17-val}} & \multicolumn{4}{c}{\textbf{MOT20-val}} \\
\cline{2-9}
\multirow{2}{*}{Distance} & \multicolumn{2}{c|}{full Hz} & \multicolumn{2}{c|}{1 Hz} & \multicolumn{2}{c|}{full Hz} & \multicolumn{2}{c}{1 Hz} \\
\cline{2-9}
& HOTA & AssA & HOTA & AssA & HOTA & AssA & HOTA & AssA \\
\noalign{\smallskip}
\cline{1-9}
\noalign{\smallskip}
Mahalanobis
& $69.4$ & $73.3$ & $55.2$ & $49.8$ & $68.0$ & $68.9$ & $50.5$ & $39.0$ \\
\hline
\rowcolor{gray!10}
BBD
& $\mathbf{69.6}$ & $\mathbf{73.7}$ & $\mathbf{64.9}$ & $\mathbf{68.2}$ & $\mathbf{68.1}$ & $\mathbf{69.0}$ & $\mathbf{63.5}$ & $\mathbf{61.5}$ \\
\hline
\end{tabular}
\caption{Comparison of the Mahalanobis distance with our Bbox-Based Distance (BBD) on different detection frequency.}
\label{table:abl-mah-bbd}
\end{table}
In this study (\cref{table:abl-mah-bbd}), we present a result of a direct comparison of our Bbox-Based Distance with the classical Mahalanobis distance across both protocols within our StableTrack pipeline. The results demonstrate that BBD performs on par with the Mahalanobis distance in the full-frequency setting. However, in the low-frequency scenario, BBD provides a significant advantage, improving HOTA by $9.7\%$ on MOT17-val and $13\%$ on MOT20-val. This validates our hypothesis that BBD, by leveraging both spatial and temporal information independent of the Kalman Filter covariance, is a more robust similarity measure in case of low-frequency detections.

\subsubsection{Role of Backward Visual Tracking}

\begin{table}
\centering
\small
\fontsize{8pt}{10pt}\selectfont
\setlength{\tabcolsep}{2pt}
\renewcommand{\arraystretch}{1.0}
\begin{tabular}{ll|cc|cc}
\hline
\noalign{\smallskip}
& & \multicolumn{2}{c|}{\textbf{MOT17-val}} & \multicolumn{2}{c}{\textbf{MOT20-val}} \\
\cline{1-6}
FVT & BVT & HOTA & AssA & HOTA & AssA \\
\noalign{\smallskip}
\cline{1-6}
\noalign{\smallskip}
\multicolumn{1}{c}{\ding{51}} & & $63.4$ & $65.1$ & $62.6$ & $59.6$ \\
\hline
\rowcolor{gray!10}
\multicolumn{1}{c}{\ding{51}} & \multicolumn{1}{c|}{\ding{51}} & $\mathbf{64.9}$ & $\mathbf{68.2}$ & $\mathbf{63.5}$ & $\mathbf{61.5}$ \\
\hline
\end{tabular}
\caption{Ablation study for Forward visual tracking (FVT) and Backward visual tracking (BVT) in case of low-frequency detections (1 Hz).}
\label{table:abl-vt}
\end{table}
In this study (\cref{table:abl-vt}), we investigate the critical role of intermediate frame utilization for ensuring accurate visual tracking performance. We compare against a baseline that uses only Forward VT~(first row) to propogate tracklets directly in the current frame $I^t$~(i.e., a full $\Delta t$ step ahead). The results show a clear advantage for our approach that uses both modules~(second row) to predict tracklets and detections positions in the intermediate frame $I^{t - \Delta t/2}$. This performance gap empirically confirms that the accuracy of visual trackers degrades over longer time intervals between frames. 

\subsection{Computational Cost}
\label{comp}
The computational efficiency of StableTrack is evaluated on the MOT17-val, with end-to-end pipeline latency measured across all components --- including object detection, feature extraction, and (visual) tracking --- under resource-constrained conditions (see \cref{subsec:impl}). Experimental results indicate a system latency of $207.5$ ms, demonstrating that our method is suitable for real-time deployment at detection frequencies of $4$ Hz and below. Additional computational analysis is provided in the supplementary materials.

\section{Conclusion}
\label{sec:conclusion}
In this paper, we introduced StableTrack, a novel multi-object tracking method designed to address the critical challenge of low-frequency detections within the tracking-by-detection framework. Motivated by the limitations of existing methods, we proposed three key contributions: (1) the robust two-stage matching strategy that improves the cross-frame association between low-frequency detections; (2) the Bbox-Based Distance (BBD), a new similarity measure that effectively incorporates spatial and temporal information; and (3) the integration of a visual tracking model that improves the Kalman Filter predictions and the overall tracking pipeline. Extensive experimental evaluation demonstrated that our method achieves state-of-the-art performance in case of low-frequency detections on MOT17, MOT20 and DanceTrack datasets, significantly outperforming existing methods, confirming its general robustness and practicality for real-world applications. Crucially, StableTrack maintains competitive performance on standard full-frequency benchmarks.
{
    \small
    \bibliographystyle{ieeenat_fullname}
    \bibliography{main}
}

\clearpage
\setcounter{page}{1}
\appendix
\maketitlesupplementary
\section{Kalman Filter}
\addcontentsline{toc}{section}{Kalman Filter}
The Kalman Filter is used to predict the position of tracks on the current frame. The Kalman Filter consists of prediction and update steps. It can be summarized by the following recursive equations:
\begin{enumerate}
\item \textbf{Prediction}:
\begin{align}
\label{KF_pre}
\hat{\mathbf{x}}_{k \mid k-1}&=\mathbf{F}_{k} \hat{\mathbf{x}}_{k-1 \mid k-1} \\
\mathbf{P}_{k \mid k-1}&=\mathbf{F}_{k} \mathbf{P}_{k-1 \mid k-1} \mathbf{F}_{k}^{\top}+\mathbf{Q}_{k}
\end{align}

\item \textbf{Update}:
\begin{align}
&\mathbf{K}_{k} = \mathbf{P}_{k \mid k-1}\mathbf{H}_{k}^{\top}(\mathbf{H}_{k}\mathbf{P}_{k \mid k-1}\mathbf{H}_{k}^{\top}+\mathbf{R}_{k})^{-1}\\
&\hat{\mathbf{x}}_{k \mid k}=\hat{\mathbf{x}}_{k \mid k-1}+\mathbf{K}_{k}(\mathbf{z}_k-\mathbf{H}_{k}\hat{\mathbf{x}}_{k \mid k-1})\\
&\mathbf{P}_{k \mid k}=(\mathbf{I}-\mathbf{K}_{k}\mathbf{H}_{k})\mathbf{P}_{k \mid k-1}
\end{align}
\end{enumerate}
At each step $k$, Kalman Filter predicts the state vector $\hat{\mathbf{x}}_{k \mid k-1}$ and the covariance matrix $\mathbf{P}_{k \mid k-1}$. Kalman Filter updates the state vector $\hat{\mathbf{x}}_{k \mid k}$ given the observation $\mathbf{z}_k$ and the estimated covariance $\mathbf{P}_{k \mid k}$ , calculated based on the optimal Kalman gain $\mathbf{K}_{k}$.

In StableTrack, the observation vector is extended to 6 dimensions $\hat{\mathbf{x}} =[x_c , y_c , w, h, v_x , v_y]^T$ , where $(x_c, y_c)$ are the 2D coordinates of the object's center, $(w, h)$ are the width and height of the object and $(v_x, v_y)$ is the estimated by Forward VT object's velocity. In order to observation vector extension, the corresponding observation matrix for the Kalman Filter is as follows:
\begin{align}
\label{KF_H}
    \mathbf{H}_k=
    \begin{bmatrix}
    1 & 0 & 0 & 0 & 0 & 0 & 0 & 0\\
    0 & 1 & 0 & 0 & 0 & 0 & 0 & 0\\
    0 & 0 & 1 & 0 & 0 & 0 & 0 & 0\\
    0 & 0 & 0 & 1 & 0 & 0 & 0 & 0\\
    0 & 0 & 0 & 0 & 1 & 0 & 0 & 0\\
    0 & 0 & 0 & 0 & 0 & 1 & 0 & 0\\
    \end{bmatrix}
\end{align}
The state transition matrix $\mathbf{F}_{k}$ from \cref{KF_pre} propagates the system state according to the constant velocity model and is defined as:
\begin{align}
\label{KF_F}
    \mathbf{F}_k=
    \begin{bmatrix}
    1 & 0 & 0 & 0 & 1 & 0 & 0 & 0\\
    0 & 1 & 0 & 0 & 0 & 1 & 0 & 0\\
    0 & 0 & 1 & 0 & 0 & 0 & 1 & 0\\
    0 & 0 & 0 & 1 & 0 & 0 & 0 & 1\\
    0 & 0 & 0 & 0 & 1 & 0 & 0 & 0\\
    0 & 0 & 0 & 0 & 0 & 1 & 0 & 0\\
    0 & 0 & 0 & 0 & 0 & 0 & 1 & 0\\
    0 & 0 & 0 & 0 & 0 & 0 & 0 & 1\\
    \end{bmatrix}
\end{align}
Notably, the object's velocity $(v_x, v_y)$ in StableTrack is explicitly estimated through the Forward VT module, rather than being modeled as a latent state parameter within the Kalman filter. All other matrices of the Kalman Filter have the same parameter values as in~\cite{tracktrack}.
\section{Bbox-Based Distance}
\addcontentsline{toc}{section}{Bbox-Based Distance}
\begin{table*}
\centering
\begin{subtable}[b]{0.32\linewidth}
\centering
\small
\setlength{\tabcolsep}{3.0pt}
\renewcommand{\arraystretch}{1.2}
\begin{tabular}{l|cc}
\hline
\noalign{\smallskip}
$\alpha$ & HOTA & AssA \\
\noalign{\smallskip}
\cline{1-3}
\noalign{\smallskip}
$0.005$ & $69.5$ & $73.5$\\
$0.015$ & $69.6$ & $73.6$\\
\hline
\rowcolor{gray!10}
$0.025$ & $\mathbf{69.6}$ & $\mathbf{73.7}$\\
\hline
$0.035$ & $69.5$ & $73.5$\\
$0.045$ & $69.4$ & $73.3$\\
 \hline
\end{tabular}
\label{table:suppl-abl-alpha}
\end{subtable}
\begin{subtable}[b]{0.32\linewidth}
\centering
\small
\setlength{\tabcolsep}{3.0pt}
\renewcommand{\arraystretch}{1.2}
\begin{tabular}{l|cc}
\hline
\noalign{\smallskip}
$\beta$ & HOTA & AssA \\
\noalign{\smallskip}
\cline{1-3}
\noalign{\smallskip}
$0.05$ & $69.5$ & $73.5$\\
$0.15$ & $69.6$ & $73.7$\\
\hline
\rowcolor{gray!10}
$0.25$ & $\mathbf{69.6}$ & $\mathbf{73.7}$\\
\hline
$0.35$ & $69.6$ & $73.6$\\
$0.45$ & $69.6$ & $73.6$\\
 \hline
\end{tabular}
\label{table:suppl-abl-beta}
\end{subtable}
\begin{subtable}[b]{0.32\linewidth}
\centering
\small
\setlength{\tabcolsep}{3.0pt}
\renewcommand{\arraystretch}{1.2}
\begin{tabular}{l|cc}
\hline
\noalign{\smallskip}
$c$ & HOTA & AssA \\
\noalign{\smallskip}
\cline{1-3}
\noalign{\smallskip}
$0.8$ & $69.5$ & $73.5$\\
$0.9$ & $69.5$ & $73.5$\\
\hline
\rowcolor{gray!10}
$1.0$ & $\mathbf{69.6}$ & $\mathbf{73.7}$\\
\hline
$1.1$ & $69.6$ & $73.6$\\
$1.2$ & $69.4$ & $73.2$\\
\hline
\end{tabular}
\label{table:suppl-abl-c}
\end{subtable}
\caption{Ablation study for $\alpha$, $\beta$ and $c$ on MOT17-val.}
\label{tab:suppl-abl-bbd}
\end{table*}
The Bbox-Based Distance (BBD) computation employs a covariance matrix $P$, which is defined as:
\begin{equation}
P = \begin{bmatrix}
    (c \cdot w)^2 \cdot [\Delta\tau]_\alpha^\beta & 0 \\
    0 & (c \cdot h)^2 \cdot [\Delta\tau]_\alpha^\beta
    \end{bmatrix},
\end{equation}
where
\begin{align*}
[\Delta\tau]_\alpha^\beta = \min(\max(\Delta \tau, \alpha), \beta),
\end{align*}
$h$ and $w$ represent the height and width from the state vector of Kalman Filter, $\Delta \tau$ is the time interval since the last successful update of the tracklet, $\alpha$, $\beta$ and $c$ are the clipping constants. Parameters $\alpha$ and $\beta$ establish the lower and upper bounds for the components of the covariance matrix. Parameter $c$ modulates the rate of variation in these components as functions of the bounding box dimensions $h$ and $w$. In StableTrack, the $\alpha$ is set to $0.025$, the $\beta$ is set to $0.25$, and the $c$ is set to $1.0$. This is justified by the results of the experiments presented in \cref{tab:suppl-abl-bbd}.
\section{Cross-Frame Association}
\addcontentsline{toc}{section}{Cross-Frame Association}
\begin{table*}
\centering
\begin{subtable}[b]{0.245\linewidth}
\centering
\small
\setlength{\tabcolsep}{3.0pt}
\renewcommand{\arraystretch}{1.2}
\begin{tabular}{l|cc}
\hline
\noalign{\smallskip}
$\theta_{\text{BBD}}$ & HOTA & AssA \\
\noalign{\smallskip}
\cline{1-3}
\noalign{\smallskip}
$8$ & $68.9$ & $72.2$\\
$12$ & $69.3$ & $73.1$\\
\hline
\rowcolor{gray!10}
 $16$ & $\mathbf{69.6}$ & $\mathbf{73.7}$\\
\hline
$20$ & $69.1$ & $72.7$\\
$24$ & $68.7$ & $71.9$\\
 \hline
\end{tabular}
\label{table:suppl-abl-bbd-thr}
\end{subtable}
\begin{subtable}[b]{0.245\linewidth}
\centering
\small
\setlength{\tabcolsep}{3.0pt}
\renewcommand{\arraystretch}{1.2}
\begin{tabular}{l|cc}
\hline
\noalign{\smallskip}
$\theta_{\text{IoU}}$ & HOTA & AssA \\
\noalign{\smallskip}
\cline{1-3}
\noalign{\smallskip}
$0.2$ & $69.4$ & $73.3$\\
$0.3$ & $69.5$ & $73.4$\\
\hline
\rowcolor{gray!10}
$0.4$ & $\mathbf{69.6}$ & $\mathbf{73.7}$\\
\hline
$0.5$ & $69.4$ & $73.3$\\
$0.6$ & $68.7$ & $71.9$\\
 \hline
\end{tabular}
\label{table:suppl-abl-iou-thr}
\end{subtable}
\begin{subtable}[b]{0.245\linewidth}
\centering
\small
\setlength{\tabcolsep}{3.0pt}
\renewcommand{\arraystretch}{1.2}
\begin{tabular}{l|cc}
\hline
\noalign{\smallskip}
$\theta_{\text{reid-high}}$ & HOTA & AssA \\
\noalign{\smallskip}
\cline{1-3}
\noalign{\smallskip}
$0.45$ & $68.6$ & $71.7$\\
$0.55$ & $69.1$ & $72.6$\\
\hline
\rowcolor{gray!10}
$0.65$ & $\mathbf{69.6}$ & $\mathbf{73.7}$\\
\hline
$0.75$ & $68.8$ & $72.2$\\
$0.85$ & $68.4$ & $71.5$\\
 \hline
\end{tabular}
\label{table:suppl-abl-reid-high}
\end{subtable}
\begin{subtable}[b]{0.245\linewidth}
\centering
\small
\setlength{\tabcolsep}{3.0pt}
\renewcommand{\arraystretch}{1.2}
\begin{tabular}{l|cc}
\hline
\noalign{\smallskip}
$\theta_{\text{reid-low}}$ & HOTA & AssA \\
\noalign{\smallskip}
\cline{1-3}
\noalign{\smallskip}
$0.1$ & $69.1$ & $72.8$\\
$0.2$ & $69.5$ & $73.5$\\
\hline
\rowcolor{gray!10}
$0.3$ & $\mathbf{69.6}$ & $\mathbf{73.7}$\\
\hline
$0.4$ & $68.9$ & $72.4$\\
$0.5$ & $68.6$ & $71.8$\\
\hline
\end{tabular}
\label{table:suppl-abl-reid-low}
\end{subtable}
\caption{Ablation study for $\theta_{\text{BBD}}$, $\theta_{\text{IoU}}$, $\theta_{\text{reid-high}}$ and $\theta_{\text{reid-low}}$ on MOT17-val.}
\label{tab:suppl-abl-tsm}
\end{table*}
During the first cross-frame association stage, the Bbox-Based Distance (BBD) between tracklets and detections is computed within the intermediate frame representation. Tracklet-detection pairs with BBD values exceeding the threshold $\theta_{\text{BBD}}$ are discarded from subsequent consideration. The second association stage implements analogous filtering using IoU with threshold $\theta_{\text{IoU}}$. Both stages also incorporate Re-ID similarity filtering, employing $\theta_{\text{reid-high}}$ threshold for the first stage and $\theta_{\text{reid-low}}$ threshold for the second stage. In StableTrack, the $\theta_{\text{BBD}}$ is set to $16$, the $\theta_{\text{IoU}}$ is set to $0.4$, and the $\theta_{\text{reid-high}}$ and $\theta_{\text{reid-low}}$ are set to $0.65$ and $0.3$, respectively. This is justified by the results of the experiments presented in \cref{tab:suppl-abl-tsm}.
\section{Qualitative Results}
\addcontentsline{toc}{section}{Qualitative Results}
\begin{figure*}[htbp]
    \centering
    \begin{tabular}{cc}
    \begin{subfigure}{0.45\textwidth}
        \centering
        \includegraphics[width=\linewidth]{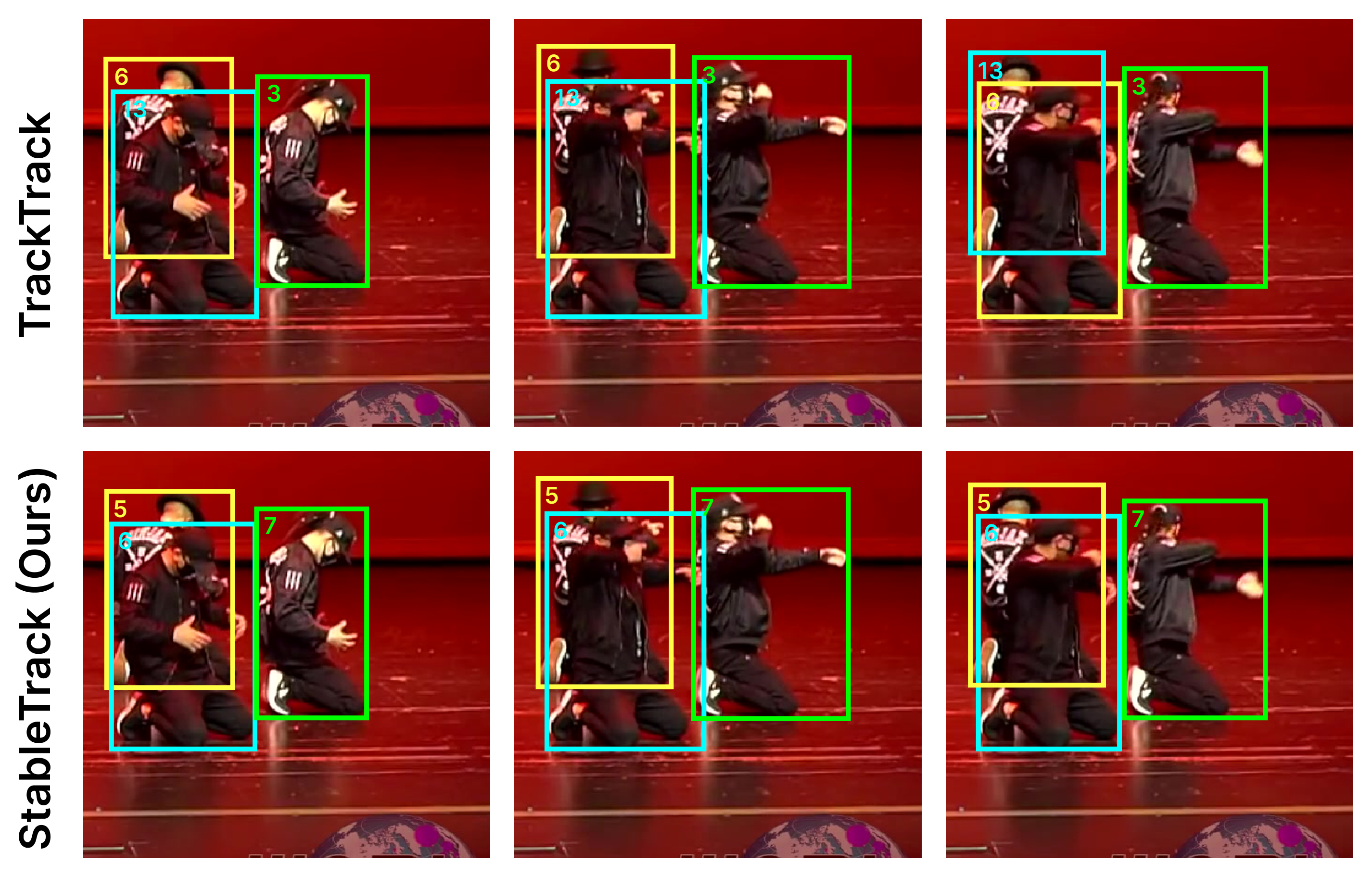}
        \caption{}
        \label{fig:a}
    \end{subfigure} &
    \begin{subfigure}{0.45\textwidth}
        \centering
        \includegraphics[width=\linewidth]{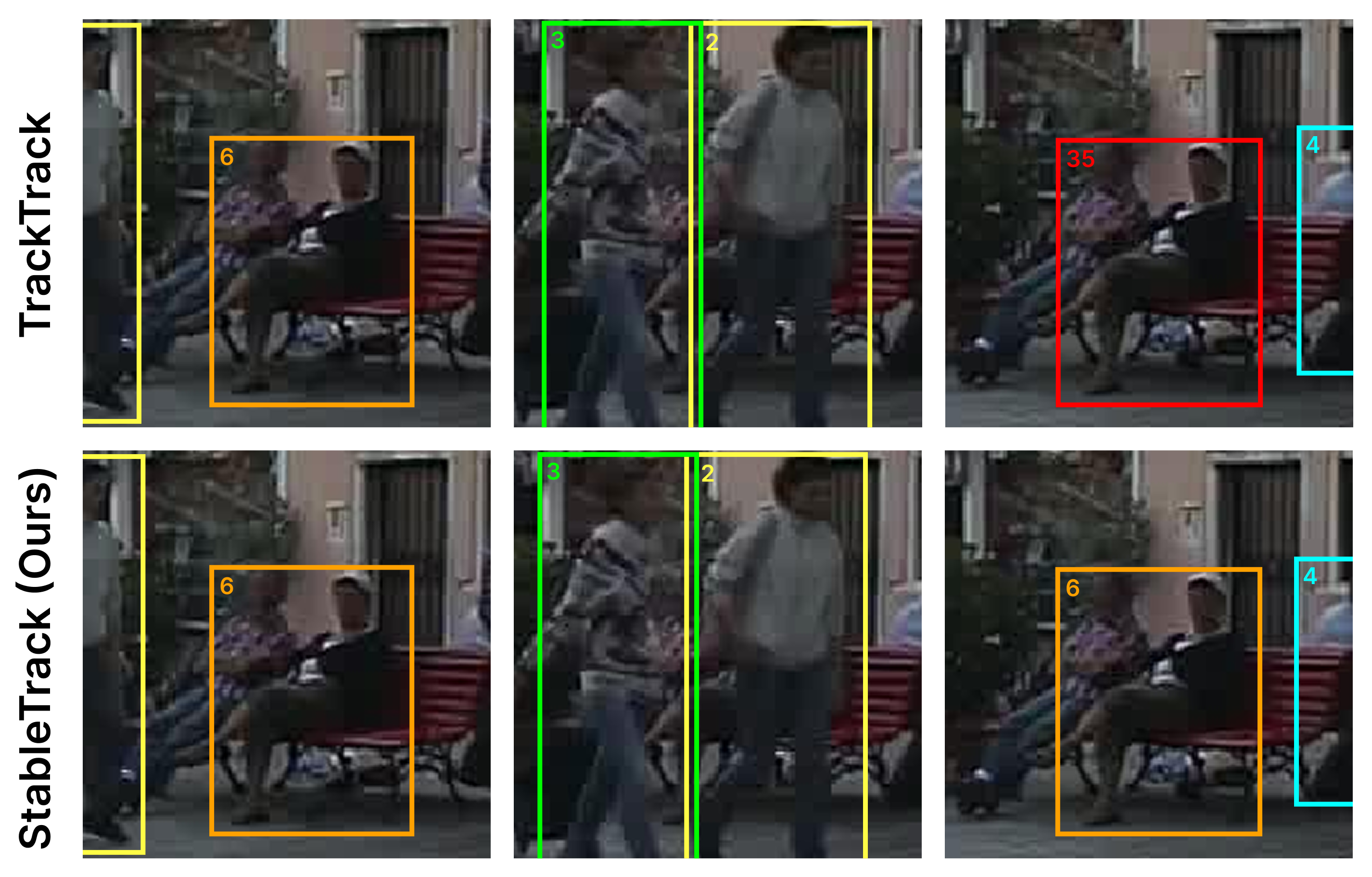}
        \caption{}
        \label{fig:b}
    \end{subfigure} \\
    
    \begin{subfigure}{0.45\textwidth}
        \centering
        \includegraphics[width=\linewidth]{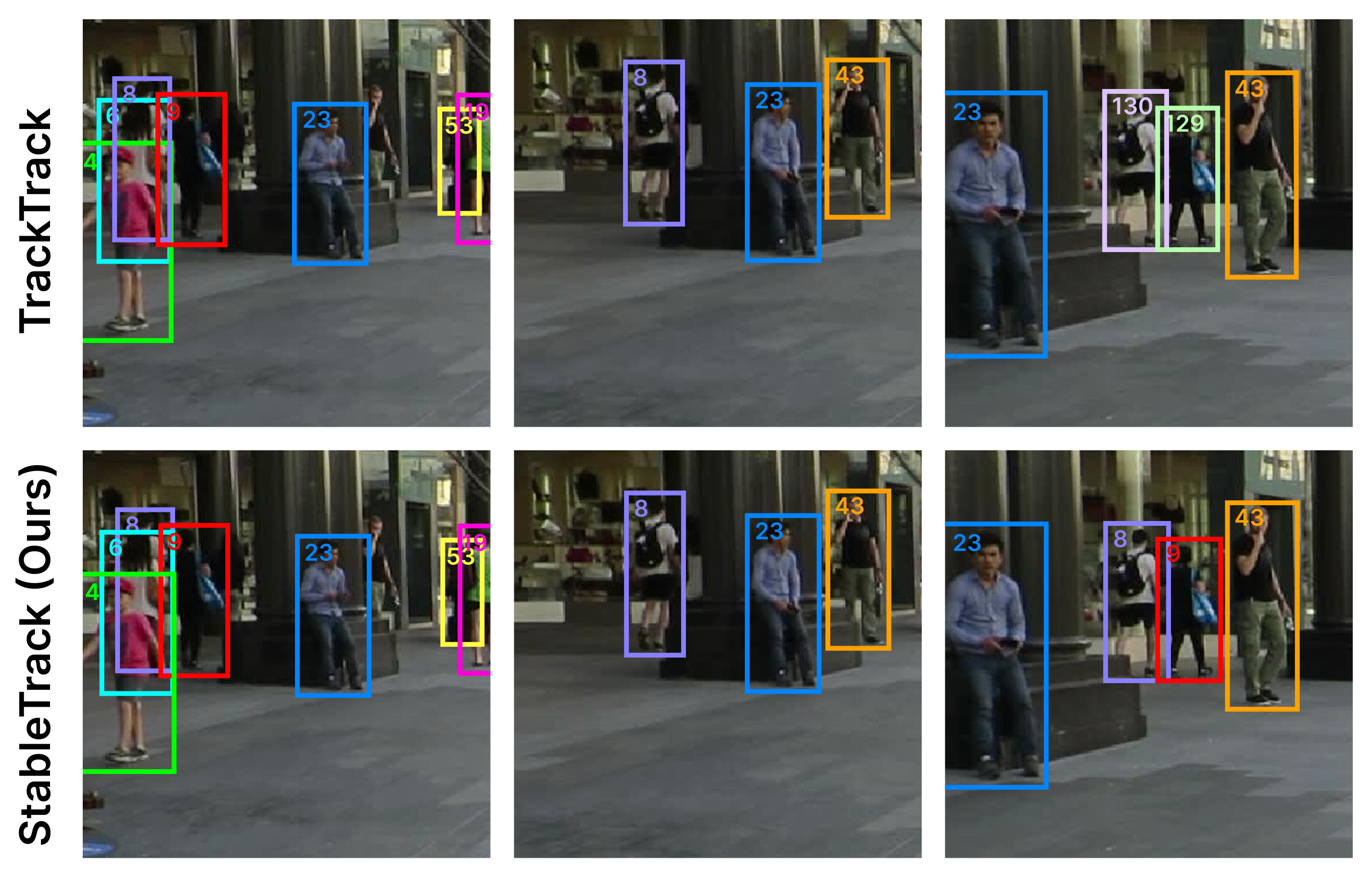}
        \caption{}
        \label{fig:c}
    \end{subfigure} &
    \begin{subfigure}{0.45\textwidth}
        \centering
        \includegraphics[width=\linewidth]{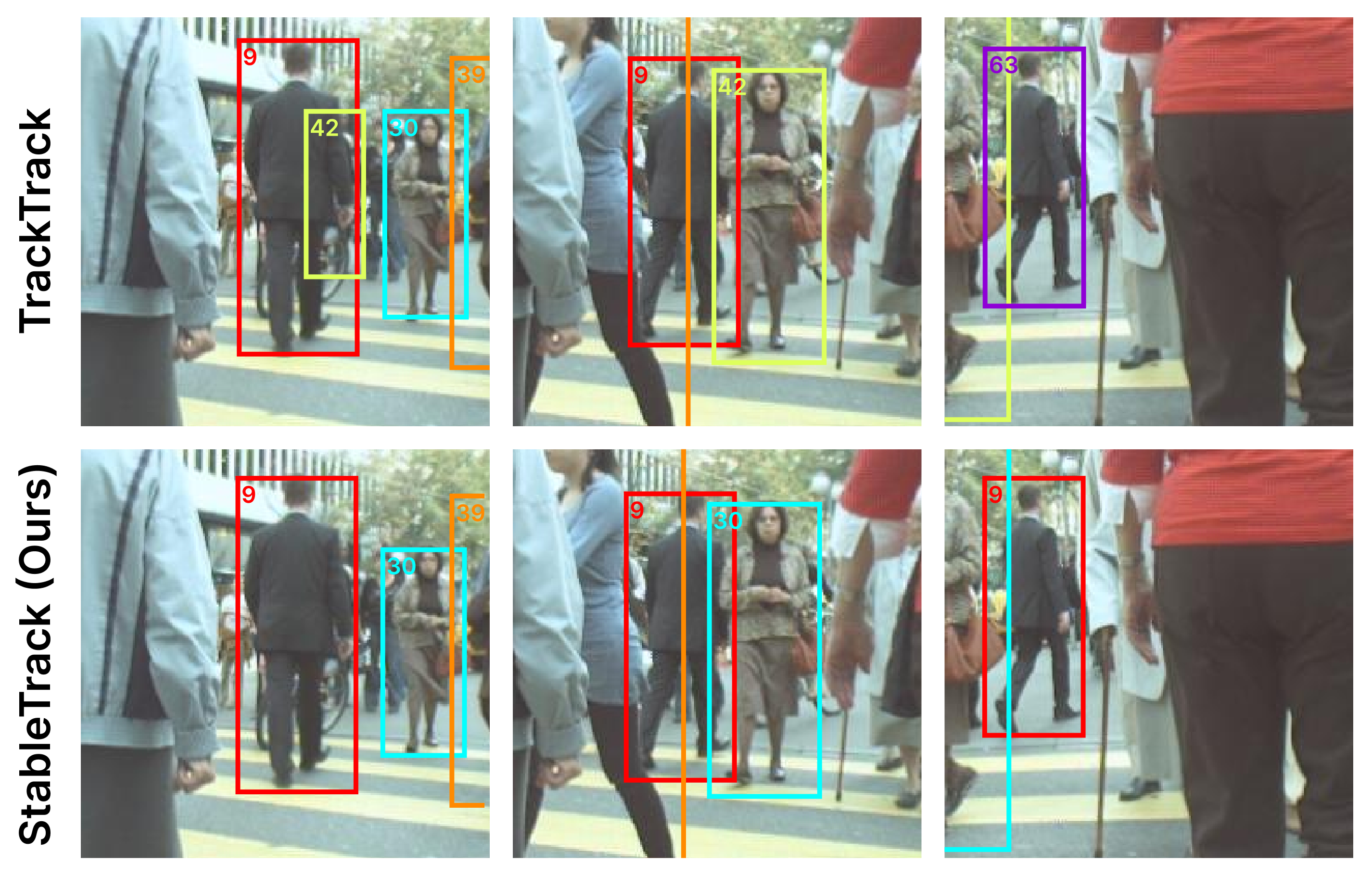}
        \caption{}
        \label{fig:d}
    \end{subfigure} \\
    \end{tabular}
    \caption{Qualitative comparisons of our StableTrack with the baseline --- current state-of-the-art method TrackTrack~\cite{tracktrack}. In the results from the baseline tracker, (a) ID $6$ and $13$ are switched because of high IoU, (b) ID $6$ is changed to $35$ after occlusion, (c) ID $8$ and $9$ are changed after the long occlusion, and (d) ID $9$ and $30$ are changed after the camera motion. In contrast, our TrackTrack shows correct tracking results in every case, demonstrating its robustness.}
    \label{fig:qual}
\end{figure*}

\begin{figure*}[htbp]
    \centering
    \begin{tabular}{cc}
    \begin{subfigure}{0.6\textwidth}
        \centering
        \includegraphics[width=\linewidth]{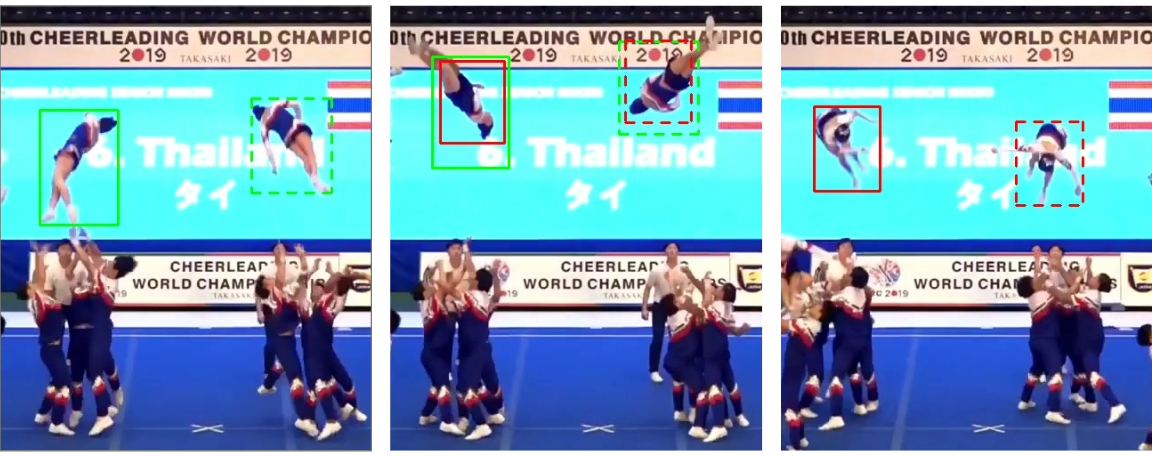}
        \caption{}
        \label{fig:asms-a}
    \end{subfigure} \\

    \begin{subfigure}{0.6\textwidth}
        \centering
        \includegraphics[width=\linewidth]{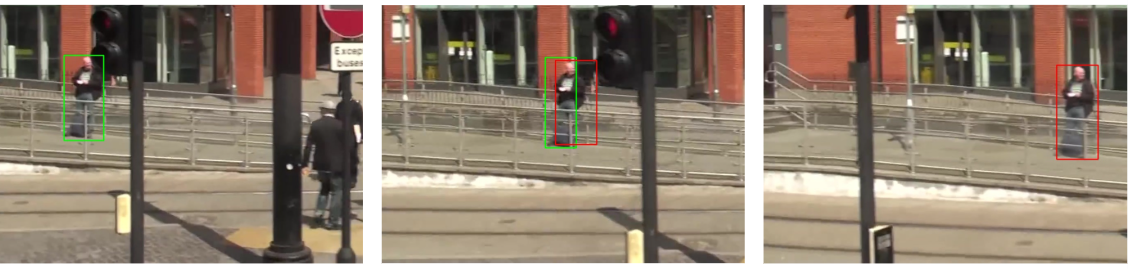}
        \caption{}
        \label{fig:asms-b}
    \end{subfigure}
    \end{tabular}
    \caption{Examples of ASMS~\cite{asms} predictions in challenging scenarios.}
    \label{fig:qual-asms}
\end{figure*}
\cref{fig:qual} presents a comparative visualization of our StableTrack with the current state-of-the-art method TrackTrack~\cite{tracktrack} under low-frequency detection conditions.

\cref{fig:a} illustrates synchronized motion patterns where persons with ID $5$ and $6$ simultaneously raise their arms, creating ambiguous overlapping regions that challenge conventional IoU-based association. By prioritizing Re-ID similarity within our cross-frame matching strategy, StableTrack effectively prevents identity switches and maintains consistent tracking despite the increased spatial ambiguity.

\cref{fig:b} demonstrates an occlusion scenario where ID $6$ becomes temporarily obscured. While the baseline method fails to re-identify the same individual, our method preserves the correct identity throughout the occlusion period.

In \cref{fig:c}, extended occlusion tests reveal StableTrack robustness in handling prolonged disappearance intervals. The temporal modeling inherent in our BBD formulation enables successful identity recovery for persons with ID $8$ and ID $9$ after substantial occlusion durations.

In \cref{fig:d}, significant camera motion induces substantial displacement of pedestrian positions (ID $9$ and $30$). Our method successfully maintains trajectory continuity through visual tracking-based position estimation, correctly recovering all original identities despite the abrupt viewpoint change.

\cref{fig:qual-asms} presents a qualitative analysis of visual tracking performance, comprising two scenarios (\cref{fig:asms-a} and \cref{fig:asms-b}). Each scenario displays three frames from two different videos. The first and third frames of each subfigure represent detection frames, on which the detector and tracking pipeline are run under the low-frequency detections, while the intermediate frame (the second) serves as a supplemental frame for predicting the position of objects using the Forward VT and Backward VT modules. In both subfigures, green bboxes represent objects, detected in the first frame, and their corresponding positions in the intermediate frame, predicted by Forward VT. Red bboxes represent objects, detected in the third frame, and their corresponding positions in the intermediate frame, predicted by Backward~VT.

In \cref{fig:asms-a}, a challenging tracking scenario demonstrates substantial inter-frame displacement of two objects with divergent velocity profiles and trajectory patterns. The visual tracking successfully maintains precise object localization in the intermediate frame, enabling robust correspondence establishment during subsequent cross-frame association stage.

\cref{fig:asms-b} illustrates a scenario with significant camera movement, causing the subject to move despite it actually being stationary. Here, the Forward VT and Backward VT modules achieve accurate position interpolation, whereas standalone Kalman Filter cannot account for possible camera movement.
\section{Impact of Re-ID Model}
\addcontentsline{toc}{section}{Impact of Re-ID Model}
\begin{table}
\centering
\small
\setlength{\tabcolsep}{2pt}
\renewcommand{\arraystretch}{1.2}
\begin{tabular}{l|cc|cc}
\hline
\noalign{\smallskip}
 \multirow{2}{*}{Re-ID} & \multicolumn{2}{c|}{full Hz} & \multicolumn{2}{c}{$1$ Hz} \\
\cline{2-5}
 & HOTA & AssA & HOTA & AssA \\
\noalign{\smallskip}
\cline{1-5}
\noalign{\smallskip}
SBS50~\cite{sbs50} & $67.7$ & $68.3$ & $56.3$ & $48.6$ \\
\hline
\rowcolor{gray!10}
DynaMix~\cite{dynamix}& $\mathbf{68.1}$ & $\mathbf{69.0}$ & $\mathbf{63.5}$ & $\mathbf{61.5}$ \\
\hline
\end{tabular}
\caption{Comparison of the conventional SBS50~\cite{sbs50} with generalized DynaMix model~\cite{dynamix} on different detection frequency on MOT20-val.}
\label{table:abl-reid}
\end{table}
In this study (\cref{table:abl-reid}), we provide a comparison of the conventional Re-ID model SBS50~\cite{sbs50}, commonly used in state-of-the-art methods~\cite{botsort, strongsort, deepocsort, hybridsort, ucmc, tracktrack}, with the generalized DynaMix model~\cite{dynamix}, used in StableTrack. For full-frequency detections scenario, the selection of Re-ID models exhibits minimal influence on overall tracking performance. Conversely, under low-frequency detection conditions, employing a generalized Re-ID architecture yields substantial improvements in tracking accuracy. Notably, even when utilizing a conventional SBS50 Re-ID model, our method maintains superior performance compared to other state-of-the-art methods in case of low-frequency detections~(see the main paper), thereby further validating the efficacy of the proposed components.
\section{Impact of Visual Tracking Model}
\addcontentsline{toc}{section}{Impact of Visual Tracking Model}
\begin{table}
\centering
\small
\setlength{\tabcolsep}{2pt}
\renewcommand{\arraystretch}{1.2}
\begin{tabular}{l|ccc}
\hline
\noalign{\smallskip}
 \multirow{2}{*}{Visual Tracking} & \multicolumn{3}{c}{$1$ Hz} \\
\cline{2-4}
 & HOTA & AssA & FPS\\
\noalign{\smallskip}
\cline{1-4}
\noalign{\smallskip}
ECO~\cite{eco} & $\mathbf{65.4}$ & $\mathbf{69.1}$ & $0.8$ \\
Staple~\cite{staple} & $65.0$ & $68.3$ & $12.8$ \\
\hline
\rowcolor{gray!10}
ASMS~\cite{asms} & $64.9$ & $68.2$ & $\mathbf{14.7}$ \\
\hline
\end{tabular}
\caption{Comparison of visual tracking models integrated within the StableTrack on MOT17-val.}
\label{table:abl-vt-suppl}
\end{table}
\cref{table:abl-vt-suppl} presents a comparative analysis of visual tracking models integrated within the StableTrack, evaluated on the MOT17-val set under a low-frequency detection protocol. Computational performance of each visual tracker was measured within the entire pipeline, specifically during execution in both Forward and Backward VT modules. Empirical results indicate that the ECO~\cite{eco} achieves optimal tracking performance. However, the computational cost of ECO renders it impractical for real-time applications. The Staple~\cite{staple} method maintains competitive accuracy while offering substantially reduced inference time. Nevertheless, the ASMS~\cite{asms} tracker was selected for final implementation due to its minimal latency, thereby preserving real-time operational capabilities.
\section{Computational Cost}
\label{comp-suppl}
\begin{table}
\centering
\small
\setlength{\tabcolsep}{2pt}
\renewcommand{\arraystretch}{1.0}
\begin{tabular}{l|c}
\hline
\noalign{\smallskip}
Components & latency (ms) \\
\noalign{\smallskip}
\cline{1-2}
\noalign{\smallskip}
Object detection & $61.2$ \\
Feature extraction & $71.7$ \\
Tracking & $74.6$ \\
\hline
\rowcolor{gray!10}
Overall & $207.5$ \\
\hline
\end{tabular}
\caption{Computational costs of a full pipeline, including object detection, feature extraction, and tracking, of our StableTrack at each iteration on MOT17-val.}
\label{table:table-comp}
\end{table}
\cref{table:table-comp} summarizes the computational efficiency of our StableTrack on the MOT17-val, reporting end-to-end pipeline latency across all components --- object detection, feature extraction, and (visual) tracking --- under resource-constrained conditions (see the main paper). Notably, feature extraction comprises simultaneously running Re-ID and visual tracking models. Thus, this stage latency is computed as the per-frame maximum between these parallel processes. Also, as our tracking method is implemented in straightforward Python code for convenience, it has significant room for computational optimization. Nevertheless, these results demonstrate that our method is suitable for real-time deployment at detection frequencies of 4 Hz and below.

\end{document}